\documentclass[conference]{IEEEtran}
\IEEEoverridecommandlockouts

\usepackage{cite}
\usepackage{amsmath,amssymb,amsfonts}
\usepackage{algorithmic}
\usepackage{graphicx}
\usepackage{textcomp}
\usepackage{xcolor}
\usepackage{multirow}
\usepackage{svg}
\usepackage{hhline}
\usepackage{hyperref}
\hypersetup{
    colorlinks=true,
    linkcolor=blue,
    filecolor=magenta,      
    urlcolor=cyan,
    }
\usepackage[nolist,nohyperlinks]{acronym}
\def\BibTeX{{\rm B\kern-.05em{\sc i\kern-.025em b}\kern-.08em
    T\kern-.1667em\lower.7ex\hbox{E}\kern-.125emX}}

\newcommand{\C}{\mathbb{C}}
\newcommand{\R}{\mathbb{R}}

\newcommand{\done}[1]{}

\newcommand{\myvector}[1]{\begin{pmatrix} #1 \end{pmatrix}}

\DeclareMathOperator{\Cov}{Cov}

\DeclareMathOperator{\BN}{BN}
\DeclareMathOperator{\std}{std}

\begin{document}

\begin{acronym}
\acro{kan}[KAN]{Kolmogorov-Arnold Network}
\acro{ckan}[CVKAN]{Complex-Valued Kolmogorov-Arnold Network}
\acro{cvnn}[CVNN]{Complex-Valued Neural Network}
\acro{mlp}[MLP]{Multilayer Perceptron}
\acro{crbfn}[CRBFN]{Complex Radial Basis Function Neural Network}
\acro{rbf}[RBF]{Radial Basis Function}
\acro{silu}[SiLU]{Sigmoid Linear Unit}
\acro{csilu}[$\C$SiLU]{}
\acro{mae}[MAE]{Mean Average Error}
\acro{mse}[MSE]{Mean Square Error}
\acro{ce}[CE]{Cross-entropy}
\end{acronym}

\title{CVKAN: Complex-Valued Kolmogorov-Arnold Networks
\thanks{The work was supported by the Deutsche Forschungsgemeinschaft DFG (SPP2363).}
}

 \author{\IEEEauthorblockN{Matthias Wolff}
 \IEEEauthorblockA{\textit{Department of Computer Science} \\
\textit{University of Münster}\\
Münster, Germany \\
matthias.wolff@uni-muenster.de}
\and
 \IEEEauthorblockN{Florian Eilers}
 \IEEEauthorblockA{\textit{Department of Computer Science} \\
\textit{University of Münster}\\
Münster, Germany \\
florian.eilers@uni-muenster.de}
\and
 \IEEEauthorblockN{Xiaoyi Jiang}
 \IEEEauthorblockA{\textit{Department of Computer Science} \\
\textit{University of Münster}\\
Münster, Germany \\
xjiang@uni-muenster.de}
}

\maketitle

\begin{abstract}
In this work we propose \acs{ckan}, a complex-valued \acf{kan}, to join the intrinsic interpretability of \acsp{kan} and the advantages of \acp{cvnn}. We show how to transfer a \acs{kan} and the necessary associated mechanisms into the complex domain. To confirm that \acs{ckan} meets expectations we conduct experiments on symbolic complex-valued function fitting and physically meaningful formulae as well as on a more realistic dataset from knot theory. Our proposed \acs{ckan} is more stable and performs on par or better than real-valued \acsp{kan} while requiring less parameters and a shallower network architecture, making it more explainable.
\end{abstract}


\smallskip
\noindent

\textbf{\textcopyright 2025 IEEE}. Personal use of this material is permitted.  Permission from IEEE must be obtained for all other uses, in any current or future media, including reprinting/republishing this material for advertising or promotional purposes, creating new collective works, for resale or redistribution to servers or lists, or reuse of any copyrighted component of this work in other works.
The final published version (M. Wolff, F. Eilers and X. Jiang, "CVKAN: Complex-Valued Kolmogorov-Arnold Networks," 2025 International Joint Conference on Neural Networks (IJCNN), Rome, Italy, 2025, pp. 1-9, doi: 10.1109/IJCNN64981.2025.11227425) is available at IEEE Xplore:
\url{https://ieeexplore.ieee.org/document/11227425}

\section{Introduction}

The recently published \acf{kan} \cite{liu2024kan,liu2024kan20} has proven to be a successful new approach for problems especially in the field of symbolic function fitting \cite{yu2024kan} and solving physical equations \cite{wang2025kolmogorov}. Adding to its success is the intrinsic explainability of \acp{kan} because of learnable univariate functions on edges instead of linear weights and fixed nonlinear activation functions in classical \acp{mlp}. These learned univariate functions hold all model weights and can easily be visualized, thus making it intuitively understandable how single layers impact the model output.

Another emerging field in machine learning research are \acfp{cvnn}, which have shown great success in a multitude of applications as well as good theoretical properties in terms of overfitting and generalization \cite{lee2022complex, bassey2021survey}. Especially for applications with complex-valued input types, these models have shown promising results.

In this work we aim to combine the advantages of \acp{kan} and \acp{cvnn} within our proposed \acf{ckan}. By replacing the learnable real-valued functions on the edges of a \ac{kan} with learnable complex-valued functions we maintain the intrinsic explainability of \acp{kan} while also leveraging the direct use of complex-valued functions, making it more suitable for problems requiring complex-valued calculations.

The original \acp{kan} used fitting of B-Splines to learn the edge functions, however this training is rather slow and can be unstable, thus \cite{li2024kolmogorov} proposed to replace the originally used B-Splines with easier to learn \acp{rbf}. While it would be possible to utilize complex-valued B-Splines \cite{ahlberg1967complex, unser2000fractional, forster2006complex} to construct a complex-valued \ac{kan}, we propose to adopt the idea of \acp{rbf} to the complex domain, to benefit from their easier and faster training process.
Additionally, we propose to replace the tedious dynamic grid extension used in \acp{kan} by Batch Normalization to prevent the model to fall outside the predefined grid.

Overall, our contributions can be summarized as:
\begin{itemize}
    \item We adopt the \ac{kan} framework to the complex domain by utilizing complex-valued \acp{rbf}.
    \item We propose to add a normalization layer to efficiently solve the problem of fixed grid sizes. 
    \item We propose a framework to utilize the explainability of \acp{kan} in the complex domain.
    \item We provide our code, both for training the model and the visualization, as an open source library to promote open science. 
    \footnote{Link to code base: \url{https://github.com/M-Wolff/CVKAN}}
\end{itemize}
We evaluate our approach against real-valued baselines on three datasets and conduct ablation studies on normalization schemes as well as explainability.

\section{Related Work}
\subsection{Kolmogorov-Arnold Networks}
After the recent introduction of \acp{kan} by Liu et al. \cite{liu2024kan,liu2024kan20} there has been a great variety of attempts to apply the ideas of \acp{kan} to different fields like image processing \cite{bodner2024convolutional}, satellite image segmentation \cite{cambrin2024kan}, graph neural networks \cite{kiamari2024gkan,bresson2024kagnns} and even transformers \cite{yang2024kolmogorov}. Hou et al. \cite{hou2024comprehensive} give a detailed overview of the different applications and extensions of \acp{kan}. Their explainability can be of great value in fields where machine learning approaches are strongly regulated like survival analysis in medicine or engineering \cite{knottenbelt2024coxkan}. Multiple works have already benchmarked the performance of \acp{kan} against \acp{mlp} \cite{yu2024kan, poeta2024benchmarking} and found \acp{kan} to be a more suitable alternative in some fields. Alter et al. \cite{alter2024robustness} found that large-scale \acp{kan} are more robust against adversarial attacks as \acp{mlp} and thus form an interesting direction for further research in multiple fields.

In \cite{wang2025kolmogorov} the authors explore different partial differential equation forms based on \ac{kan} instead of \ac{mlp} for solving forward and inverse problems in computational physics. A systematical comparison demonstrates that the \ac{kan} approach significantly outperforms \ac{mlp} regarding accuracy and convergence 
speed. Further successful applications of \ac{kan} can be found for operator learning in computational mechanics \cite{Abueidda2025} and image classification \cite{Jamali2024}.

\subsection{Complex-Valued Neural Networks}
After early introduction of \acp{cvnn} \cite{aizenberg1971about, hirose2003complex, nitta1997extension, yang1994complex} they have lately risen in popularity since the introduction of building blocks for deep learning architectures \cite{trabelsi2018deep}. Since then a lot of theoretical contributions have been made \cite{tan2022real,eilers2023building,zhang2021optical} to enable a multitude of applications \cite{zhong2023real,chen2023spectral,liu2023pixelwise,xing2023phase,yakupouglu2024comparison}.

The most closely related prior work in the complex domain are deep \acp{crbfn} \cite{chen2008fully, soares2024deep}. However, in \acp{crbfn}, the \acp{rbf} are applied to all inputs of a neuron (e.g. vertex of the computational graph) simultaneously, while in \acp{ckan} the \acp{rbf} are applied on the edges of the computational graph to each value individually. Thus \acp{crbfn} are architecturally more similar to classical \acp{mlp} with \acp{rbf} as activation functions, where we instead aim to adopt the \ac{kan} framework to the complex domain.

\section{Real-Valued Kolmogorov-Arnold Networks}
 The Kolmogorov-Arnold representation theorem \eqref{eq:kan_theorem} \cite{kolmogorov1958On} states that any continuous function can be represented by a superposition of univariate functions.
 \begin{equation}
     f(x_1, \dots, x_n) = \sum_{q=1}^{2n+1} \Phi_q \left(\sum_{p=1}^{n}\phi_{q,p}(x_p) \right)
     \label{eq:kan_theorem}
 \end{equation}
 This formula can be used to construct a two layer neural network with $n$ inputs, one output and learnable functions $\Phi_q, \phi_{q,p}$. For each layer and each feature an individual univariate function has to be learned, so combination of features is only done by summing the outputs of all these functions.
 While theoretically possible, this theorem was long deemed to be of little use for machine learning \cite{girosi1989representation}, because the inner functions can be highly nonsmooth. 
 
 However, Liu et al. \cite{liu2024kan} have generalized this formulation of only one hidden layer of size $2n+1$ and one single output to arbitrary network depths $L$ and layer widths $n_l$ with $l = 0 \ldots L-1$ to overcome the problem of  nonsmooth inner functions.
 Let $N_{l,i}$ be the $i$-th neuron in layer $l$, then every node $N_{l,p}$ in layer $l$ is connected to every node $N_{l+1,q}$ in the following layer by an edge $E_{l,q,p}$ with $p=1 \dots n_l$ and $q=1 \dots n_{l+1}$. On every edge there is one learnable univariate function $\phi_{l,q,p}$ and the values of all incoming edges into a single node of the next layer are summed. This way the multiplication with linear weights between layers in classical \acp{mlp} has been replaced by learnable univariate functions and the nonlinear but fixed activation functions in the nodes of \acp{mlp} have been replaced by plain summation. In the original \ac{kan} the functions are learned using B-Splines \cite{liu2024kan, liu2024kan20}.

\subsection{Real-Valued Radial Basis Functions}
\begin{figure}[t]
    \begin{center}    
    \includegraphics[width=.8\linewidth]{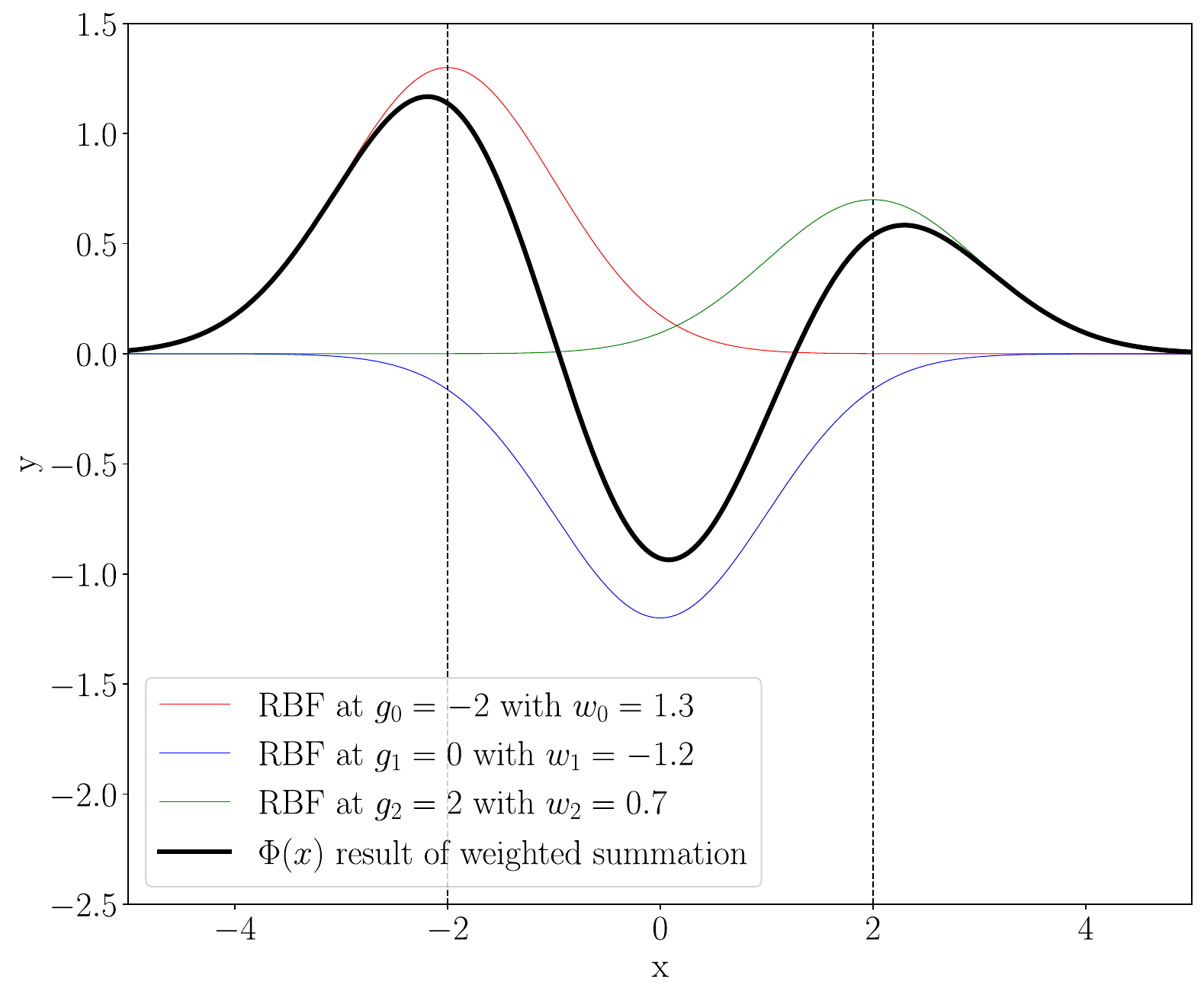}
    \caption{Visualization of a weighted sum of three \acp{rbf} with a grid in the interval $[-2, 2]$ and grid points at $g_0=-2, g_1=0, g_2=2$.}
    \label{fig:rbf_weighted_sum}
    \end{center}
\end{figure}

The way B-Splines are used to learn the functions in \acp{kan} \cite{liu2024kan,liu2024kan20} is computationally intensive and slow. Thus different improved methods for learning a function have been proposed like DeepOKAN \cite{Abueidda2025} or FastKAN \cite{li2024kolmogorov}, in which the authors suggest to replace B-Splines with \acp{rbf} for faster computation. Equation \eqref{eq:rbf} describes a \ac{rbf} that is symmetrical and centered around $0$. 
\begin{equation}
    \phi(x) = \exp \left(-x^2\right)
    \label{eq:rbf}
\end{equation}
To represent more complicated functions multiple \acp{rbf} are centered around uniformly distributed grid points $g_i$ inside a fixed grid with $i=0 \dots G-1$ and $G$ denoting the number of grid points. All those \acp{rbf} are then combined in a weighted sum with learnable weights $w_i \in \R$ to construct one continuous function over the whole grid \eqref{eq:rbf_sum}. In Fig. \ref{fig:rbf_weighted_sum} we visualize such a function made up of $G=3$ \acp{rbf}.
\begin{equation}
    \Phi (x) = \sum_{i=0}^{G-1} \left(w_i \phi(x-g_i)\right)
    \label{eq:rbf_sum}
\end{equation}
The number of grid points $G$ and the range of the grid $[a,b]$, over which the \acp{rbf} are defined, are hyperparameters. By making the grid finer one can represent more complicated functions with increasing precision.

\section{Complex-Valued Kolmogorov-Arnold Networks}
To extend \acp{kan} into the complex-valued domain we need a way to learn functions $f: \C \to \C$ on every edge. Inspired by the use of \acp{rbf} in FastKAN we propose to construct those complex functions using multiple \acp{rbf} on a 2D grid. Furthermore we employ a complex-valued equivalent to the \ac{silu} function used in \acp{kan} and make use of Batch Normalization to stay inside our fixed range grid. Finally we present a tool for visualizing the \ac{ckan} to make it interpretable and explain its inner workings.

\subsection{Complex-Valued Radial Basis Functions}
\label{sec:ckan_crbf}

Chen et al. \cite{chen2008fully} have chosen to treat real and imaginary parts separately and construct a complex-valued \ac{rbf} out of two real-valued \acp{rbf} which only depend on the real or imaginary part of the input each. On the other hand Soares et al. \cite{soares2024deep} use a single real-valued \ac{rbf} that depends on both the real and imaginary parts of the input jointly. While the latter apply these multivariate \acp{rbf} inside vertices, we propose to use univariate \acp{rbf} on edges instead to maintain the idea of \acp{kan} and achieve higher explainability.
We define a $\text{\ac{rbf}}: \C \to \R$ as:
\begin{equation}
    \phi_\C(x) = \exp\left(-|x|^2\right)
    \label{eq:crbf}
\end{equation}
We can use this \ac{rbf} in combination with complex-valued grid points and complex-valued weights to learn a complex-valued function.
Let $w_{u,v} \in \C$ be learnable weights and $g_{u,v} \in \C$ complex-valued grid points on a grid with $G \times G$ grid points to cover the real and imaginary parts of the input. Then the complex-valued activation function $\Phi_{\C}: \C \to \C$ can be represented as:
\begin{align}
    \Phi_{\C}(x) &= \sum_{u=0}^{G-1} \left( \sum_{v=0}^{G-1} w_{u,v} \hspace{0.5em} \phi_\C(x-g_{u,v}) \right) \label{eq:ckan_noskip}
\end{align} 
Note that this calculation is equivalent to learning two functions $\C\to\R$ separately with twice as many real-valued weights $\Re(w_{u,v}), \Im(w_{u,v}) \in \R$ and treating these as real- and imaginary parts of the functions learned here. 

In the special case of a real-valued output (e.g. a classification task), we drop the imaginary part of the weights in the last layer, thus obtaining a function $\C\to\R$.

\begin{figure}[t]
    \begin{center}    
    \includegraphics[width=\linewidth]{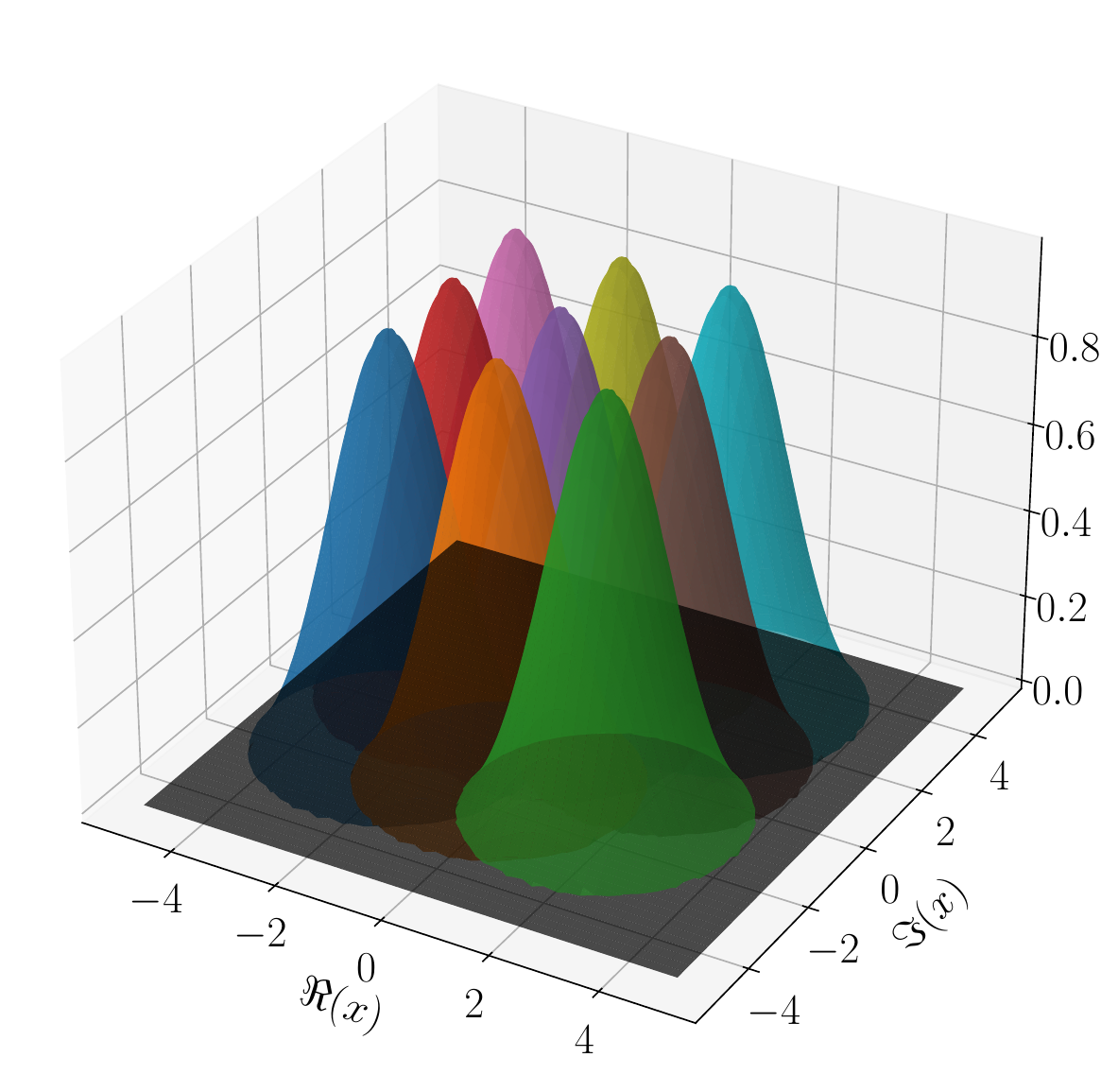}
    \caption{Visualization of the complex RBFs with a grid in the interval \mbox{$[(-2 -2i), (2+2i)]$} and $G=3$ grid points per dimension resulting in $3\cdot3=9$ grid points.}
    \label{fig:rbf_complex_cones}
    \end{center}
\end{figure}

In Fig. \ref{fig:rbf_complex_cones} we depict a visualization of the \acp{rbf} on a 2D grid. Each cone at gridpoint $g_{u,v}$, represented by a \ac{rbf} $\C \to \R$, gets multiplied by complex-valued learnable weights $w_{u,v}$ to construct a complex-valued output.

\subsection{Residual Activation Function}
\label{sec:cvkan_residual}
In the real-valued \acp{kan}, a residual activation function is used to help the training of the univariate functions. To this end an activation function $\sigma$ is added to the sum of \acp{rbf} as explained above. Thus we chose to extend \eqref{eq:ckan_noskip} to:
\begin{align}
    \Phi_\C(x) = \Phi_{\C}(x) + \sigma(x)
    \label{eq:ckan_forward_withzsilu}
\end{align}
We propose to use \acs{csilu} \eqref{eq:zsilu} as this residual activation function, a complex-valued equivalent to the \ac{silu}-function \eqref{eq:silu} used in \cite{liu2024kan,liu2024kan20,li2024kolmogorov}: 
\begin{align}
    \text{\acs{silu}}(x) &= x \left(\frac{1}{1+e^{-x}} \right)\label{eq:silu} \\
    \text{\acs{csilu}}(x)&= \text{\acs{silu}}(\Re(x)) + i \ \text{\acs{silu}}(\Im(x))
    \label{eq:zsilu}
\end{align}
This residual activation function additionally gets a learning weight and bias, which we also adopt in two ways. First, we use a complex-valued weight $w_\C \in \C$ \eqref{eq:zsiluC} on the output of the full \acs{csilu}. Second, the real and imaginary parts are each weighted with a real-valued weight $w_1, w_2 \in \R$ separately \eqref{eq:zsiluR}. Both approaches get an additive bias $\beta \in \C$. Thus the two approaches studied in our work are:
\begin{align}
    \text{\acs{csilu}}_\C(x)&= w_{\C} \left(\text{\acs{silu}}(\Re(x)) + i \ \text{\acs{silu}}(\Im(x))\right) + \beta\label{eq:zsiluC}\\
    \text{\acs{csilu}}_\R(x) &= w_1 \text{\acs{silu}}(\Re(x)) + i \ w_2 \text{\acs{silu}}(\Im(x)) + \beta \label{eq:zsiluR}
\end{align}
In our experiments in section \ref{sec:exp} we analyze the difference in performance of both approaches.

\subsection{Complex-Valued Batch Normalization}
\label{sec:cvkan_cvbn}
For \acp{kan} it is necessary to fix the grid range. However, it is not guaranteed that the output of an intermediate layer (and thus the input of the next layer) stays within a grid that was fixed before training. To tackle this challenge \cite{liu2024kan} proposed dynamic extension of the grid during training. This is however a time intensive process that might slow down the learning process. Instead we propose to normalize the output of a vertex after summation.
We propose to use BatchNorm ($\BN_\R$) \cite{ioffe2015batch} for the \acl{rbf} based FastKAN \cite{li2024kolmogorov}.
To adapt this to the complex domain, we explore three different complex-valued Batch Normalization approaches. 

First, we adapt $\C$BatchNorm \cite{trabelsi2018deep}, where the covariance matrix of the complex distribution is normalized and an output distribution is learned through a learnable covariance (e.g. a symmetric positive definite) matrix. With $\overline{z}, \Cov(z)$ being the mean and covariance over the batch dimension and $\gamma_{\Cov} \in \R^{2{\times}2}, \beta\in\C$ learnable parameters, it is then defined as:
\begin{align}
    \begin{pmatrix}
        \Re(\BN_\C(z)) \\
        \Im(\BN_\C(z))
    \end{pmatrix} = \gamma_{\Cov} \Cov(z)^{-\frac{1}{2}} \begin{pmatrix}
        \Re(z - \overline{z}) \\
        \Im(z - \overline{z})
    \end{pmatrix} + \beta
\end{align}

Second, we propose to standardize the variance of the complex-valued input distribution. With $\overline{z}, \mathbb{V}(z)$ being the mean and variance over the batch dimension and $\gamma_\R \in \R, \beta \in \C$ learnable parameters, we define it as:
\begin{align}
    \BN_{\mathbb{V}}(z) = \gamma_\C \frac{z-\overline{z}}{\sqrt{\mathbb{V}(z)}} + \beta
\end{align}

The third approach is to simply normalize the real and imaginary part of the complex-valued inputs separately. This is equivalent to just applying the real-valued BatchNorm separately to the real and imaginary parts of the input:
\begin{align}
    \BN_{\R^2}(z) = \BN_\R(\Re(z)) + i\BN_\R(\Im(z))
\end{align}
We employ these three approaches after every but the last layer in our \ac{ckan} and study their performance in section \ref{sec:exp}.

\subsection{Interpretability of Complex-Valued Kolmogorov-Arnold Networks}
One major advantage of \acp{kan} is their inherent interpretability. Since the \ac{kan} learns on the edges, the learned univariate functions are directly applied to each input feature of the current layer. As proposed in \cite{liu2024kan,liu2024kan20}, we can utilize this property in two ways to explain the inner workings of the \ac{kan}: We can give importance scores to all edges and vertices to understand which parts of the network significantly influence the output and we can visualize the learned functions on the edges to understand their input-output relation. Both of these can also be utilized for \acp{ckan}. 

To calculate importance scores for edges and vertices, \ac{kan} 2.0 \cite{liu2024kan20} uses the standard deviation of the output of that edge or vertex over the whole dataset distribution to annotate every edge and node with a relevance score by iterating backwards through the network. We adopt this idea and propose to use the standard deviation of the complex distribution. For $z=(z_1, \ldots, z_n)\in \C^n$ and $\overline{z}$ its mean, it is defined as:
\begin{align}
    \std(z) = \sqrt{\frac{1}{n-1} \sum_{i=1}^n |z_i - \overline{z}|^2}
\end{align}
Fig. \ref{fig:ckan_knot_relevances} presents an example of utilizing the standard deviation to assign relevance scores to edges and vertices as well as visualizing the most important learned edge function, i.e. the one with the highest relevance score.

For \ac{ckan}, the inputs and outputs of edges are complex-valued. To visualize their behavior we thus have to visualize a function $\C\to\C$. We propose to visualize this as a colored 3D plot, where bases are the real and imaginary part of the input, the height is the magnitude of the output and the color shows the phase of the output. Due to the $2\pi$ periodicity of the phase, we choose a periodic color map. In Fig. \ref{fig:ckan_sqsq} a small \ac{ckan} is visualized as an example.

One limitation to the interpretability in general is the depth of the network. The deeper the network, the harder it is to understand the full model by understanding its subparts. Thus, when evaluating the suitability of a \ac{kan} model, the trade-off between model size and performance maximization needs to be accounted for.

\section{Experiments}\label{sec:exp}

To compare our complex-valued \ac{ckan} with the real-valued \ac{kan} and FastKAN on complex-valued datasets we split the complex numbers of the input and output layer of real-valued \acp{kan} into real and imaginary parts. Thus we end up with real-valued \acp{kan} that have twice the input and output dimension of our \ac{ckan}.
We conduct three different experiments:
\begin{itemize}
\item Just like in \cite{liu2024kan}, we show that our network is capable of learning simple symbolic correlations between input and output based on synthetic formulae and outperforms its real-valued counterparts on complex-valued problems.
\item We extend this synthetic function-fitting task from arbitrary simple functions to more complicated and physically meaningful formulae.
\item We use the knot dataset \cite{davies2021advancing}, which contains two complex-valued features and was also used by Liu et al. to study the original \ac{kan} \cite{liu2024kan}, to analyze the performance of our \ac{ckan} on a more realistic dataset and for classification.
\end{itemize}
For our experiments based on synthetic formulae we sample data points for each variable within our grid range and calculate the target output based on the symbolic formula. Our grid range is always $[-2,2]$ in $\R$ or $[(-2, -2i), (2,2i)]$ in $\C$ respectively. Because of Batch Normalization the resulting distribution has a standard deviation of $\sigma=1$ and mean $\mu=0$. We use a grid centered around zero spanning two standard deviations in each direction to optimize the trade off between a smaller grid size and minimizing the chance of the output of the former layer exceeding the grid. For the tabular knot dataset \cite{davies2021advancing} we normalize each feature to be inside of our grid. If the dataset contains real-valued features we also feed them into our \ac{ckan} as complex-valued numbers but set the imaginary part to zero.

We always apply a 5-fold cross validation. As metrics to evaluate the performance of the models we use \ac{mae} and \ac{mse} for regression and \ac{ce}-Loss and Accuracy for classification tasks.

\subsection{Function Fitting}
\label{sec:exp_funcfit}
We chose four simple functions with one or two complex-valued input variables (cf. Table \ref{tab:funcfit_formulas}) to demonstrate that our \acs{ckan} is capable of learning basic correlations.

For each function from Table \ref{tab:funcfit_formulas} one can determine the required model size by looking at the hierarchy of the function evaluation. Since we can learn any univariate function on a single edge, for $f_1$ a $1 {\times} 1$ \acs{ckan} is optimal in theory. In the real-valued domain we need at least a $2 {\times} 3 {\times} 2$ KAN to represent the equivalent computation using only real numbers (cf. Table \ref{tab:funcfit_formulas}).
The function $f_{1\R}$ is an equivalent computation to $f_1$ using only real-valued numbers and twice as many inputs and outputs to accommodate the real and imaginary parts of the complex numbers separately. We need three neurons in the hidden layer to represent $x_1^2$, $x_2^2$ and $(x_1 + x_2)$. In the output layer we can then square the resulting value of the third hidden neuron and combine it with the (negative) identity of the first and second hidden neuron to construct the real and imaginary part of the output, each as a single real-valued number.

\begin{table}[tbp!]
\begin{center}
    \caption{Overview of the different formulae used in our function fitting experiments. Left column for the complex-valued functions and right column for their respective real-valued equivalents, with $z=x+iy$. Vector notation represents the single features that are fed into the network. Note, the outer square in $f_{4\R}$ is calculated as a multiplication in $\C$.}
    \begin{tabular}{|c|c|}
        \hline
        \rule{0pt}{5mm}Function $\C\to\C$ & Function $\R^2\to\R^2$ \\
        \hline
         \rule{0pt}{5mm}$f_1(z) = z^2$ & $f_{1\R}(x,y) = \myvector{x^2 - y^2\\2xy}$ \\
         \hline
         \rule{0pt}{5mm}$f_2(z)= \sin(z)$ & $f_{2\R}(x,y) = \myvector{\sin(x)\cosh(y)\\ \cos(x)\sinh(y)}$\\
         \hline
         \rule{0pt}{5mm}$f_3(z_1,z_2) = z_1z_2$ & $f_{3\R}(x_1,y_1,x_2,y_2) = \myvector{x_1x_2 - y_1y_2\\x_1y_2 + x_2y_1}$ \\
         \hline
         \rule{0pt}{5mm}$f_4(z_1, z_2) {=} (z_1^2+z_2^2)^2$ & $f_{4\R}(x_1, y_1, x_2, y_2){=}{\myvector{x_1^2 + x_2^2 - y_1^2 - y_2^2\\2x_1y_1 + 2x_2y_2}}^2$\\
        \hline
    \end{tabular}
    \label{tab:funcfit_formulas}
    \end{center}
\end{table}

As for the network size we chose the theoretically optimal architecture size for \ac{ckan} and regular \ac{kan} respectively and also experimented with half the size of the optimal \ac{kan} architecture for \ac{ckan} and vice versa double the optimal size of \ac{ckan} for regular \ac{kan}. The grid size is $G=8$ per dimension for \ac{ckan} and $G=8\cdot8=64$ for real-valued \ac{kan} respectively.

A total of $5000$ points were randomly sampled inside our grid for every function and models of different sizes were trained for $1000$ epochs using 5-fold cross validation.

The results of training on toy datasets generated by simple complex-valued formulae (cf. Table \ref{tab:funcfit_formulas}) can be seen in Table \ref{tab:results_ff_dummy}. For $z^2$ and $(z_1^2+z_2^2)^2$ our \ac{ckan} is by far the best performing model when compared to FastKAN and \ac{kan} with similar or even bigger numbers of parameters. For $z_1 \cdot z_2$ \ac{ckan} is slightly better than FastKAN while \ac{ckan} requires only half the number of parameters. For $\sin(z)$ our approach is slightly worse than FastKAN, however FastKAN requires a bigger network size and thus $\approx16$ times as many parameters to perform marginally better than \ac{ckan}. With \ac{ckan} we are also able to achieve on average lower standard deviations on the results of the 5-fold cross validation, emphasizing the stability of our proposed model.

 \begin{figure}[tbp!]
    \begin{center} 
    \includegraphics[width=\linewidth]{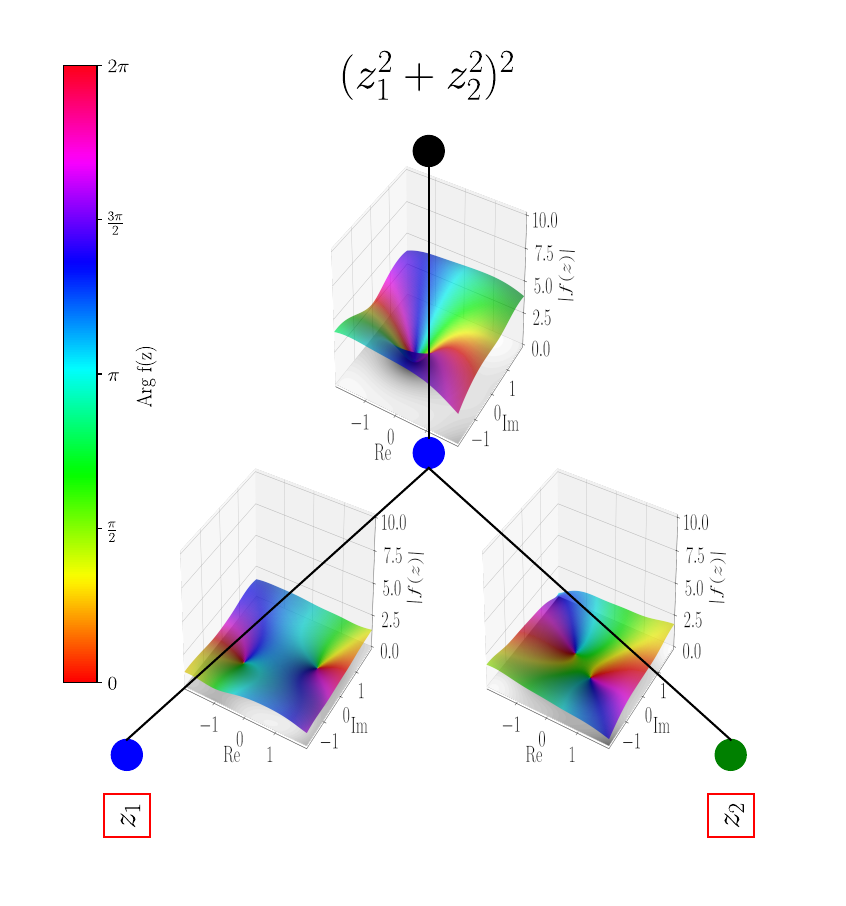}
    \caption{Visualization of a \ac{ckan} for symbolic function fitting. Height of the graphs show magnitude of the learned function, colormap encodes the phase. Both functions on the lower edges represent quadratic functions with a constant offset. They sum to $z_1^2 + z_2^2$ since their offsets cancel out, which can be seen on the roots, which are identical besides a 90° rotation. The upper function represents a typical shape of a complex square function near the middle with some error artifacts on the edges of the grid.}
    \label{fig:ckan_sqsq}
    \end{center}
\end{figure}

 \begin{figure*}[t]
    \begin{center}    
    \includegraphics[width=\linewidth]{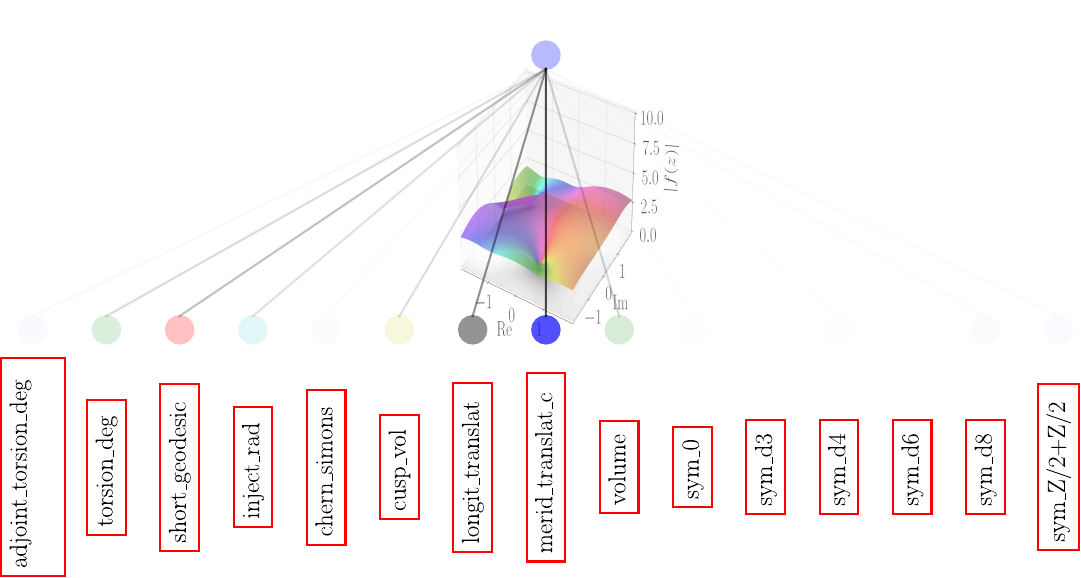}
    \caption{Visualization of the first layer of a \ac{ckan} on the knot dataset with the opacity of edges and nodes proportional to their relevance score. For better readability only the function corresponding to the most relevant blue input node for feature merid\_translat\_c is shown. In our visualization tool the individual functions and edges can be interactively selected by double-clicking on the corresponding nodes or by specifying the edge index $(l,i,j)$ to view only the function of edge $E_{l,j,i}$ in detail.}
    \label{fig:ckan_knot_relevances}
    \end{center}
\end{figure*}

\begin{table*}[tbp!]
    \begin{center}
    \caption{Results of \ac{ckan}, FastKAN and \ac{kan} on four datasets for function fitting.}
\begin{tabular}{|c|c|c|c|c|c|}
\hline
Dataset & Model & Size & \# Params & Test MSE & Test MAE \\\hline
\multirow{6}{*}{$z^2$} & \multirow{2}{*}{\acs{ckan}} & $1 {\times} 1$ & 132 & $0.014 $ \scriptsize{$\pm 0.002$} & $\mathbf{0.088}$ \scriptsize{$\pm 0.003$} \\
 & & $1 {\times} 2 {\times} 1$ & 538 & $\mathbf{0.013}$ \scriptsize{$\pm 0.004$} & $0.097 $\scriptsize{$\pm 0.016$} \\
 \cline{2-6}
 & \multirow{2}{*}{FastKAN} & $2 {\times} 2$ & 262 & $3.583 $ \scriptsize{$\pm 0.211$} & $1.006 $\scriptsize{$\pm 0.027$} \\
 &  & $2 {\times} 3 {\times} 2$ & 791 & $0.195 $ \scriptsize{$\pm 0.068$} & $0.305 $\scriptsize{$\pm 0.061$} \\
 \cline{2-6}
 & \multirow{2}{*}{KAN} & $2 {\times} 2$ & 292 & $3.482 $ \scriptsize{$\pm 0.095$} & $0.987 $\scriptsize{$\pm 0.022$} \\
 & & $2 {\times} 3 {\times} 2$ & 876 & $0.260 $ \scriptsize{$\pm 0.097$} & $0.316 $\scriptsize{$\pm 0.078$} \\
 \hline
 \multirow{6}{*}{$\sin (z)$} & \multirow{2}{*}{\acs{ckan}} & $1 {\times} 1$ & 132 & $0.005 $ \scriptsize{$\pm 0.001$} & $0.051 $\scriptsize{$\pm 0.002$} \\
 & & $1 {\times} 2 {\times} 1$ & 538 & $0.010 $ \scriptsize{$\pm 0.001$} & $0.087 $\scriptsize{$\pm 0.004$} \\
 \cline{2-6}
 & \multirow{2}{*}{FastKAN} & $2 {\times} 2$ & 262 & $0.478 $ \scriptsize{$\pm 0.029$} & $0.506 $\scriptsize{$\pm 0.016$} \\
 & & $2 {\times} 4 {\times} 4 {\times} 2$ & 2106 & $\mathbf{0.004}$ \scriptsize{$\pm 0.001$} & $\mathbf{0.047} $\scriptsize{$\pm 0.007$} \\
 \cline{2-6}
 & \multirow{2}{*}{KAN} & $2 {\times} 2$ & 292 & $0.495 $ \scriptsize{$\pm 0.021$} & $0.509 $\scriptsize{$\pm 0.010$} \\
 &  & $2 {\times} 4 {\times} 4 {\times} 2$ & 2336 & $0.363 $ \scriptsize{$\pm 0.289$} & $0.416 $\scriptsize{$\pm 0.164$} \\
\hline
\multirow{6}{*}{$z_1 \cdot z_2$} & \multirow{2}{*}{\acs{ckan}} & $2 {\times} 2 {\times} 1$ & 802 & $0.240 $ \scriptsize{$\pm 0.045$} & $0.380 $\scriptsize{$\pm 0.049$} \\
& & $2 {\times} 4 {\times} 2 {\times} 1$ & 2406 & $\mathbf{0.045}$ \scriptsize{$\pm 0.015$} & $\mathbf{0.177}$\scriptsize{$\pm 0.033$} \\
\cline{2-6}
& \multirow{2}{*}{FastKAN} & $4 {\times} 4 {\times} 2$ & 1574 & $0.769 $ \scriptsize{$\pm 0.140$} & $0.661 $\scriptsize{$\pm 0.066$} \\
 & & $4 {\times} 8 {\times} 4 {\times} 2$ & 4718 & $0.076 $ \scriptsize{$\pm 0.081$} & $0.179 $\scriptsize{$\pm 0.085$} \\
 \cline{2-6}
 & \multirow{2}{*}{KAN} & $4 {\times} 4 {\times} 2$ & 1752 & $1.779 $ \scriptsize{$\pm 0.207$} & $1.031 $\scriptsize{$\pm 0.070$} \\
 & & $4 {\times} 8 {\times} 4 {\times} 2$ & 5256 & $4.561 $ \scriptsize{$\pm 0.281$} & $1.674 $\scriptsize{$\pm 0.055$} \\
\hline
\multirow{6}{*}{$(z_1^2 + z_2^2)^2$} & \multirow{2}{*}{\acs{ckan}} & $2 {\times} 1 {\times} 1$ & 401 & $320.232 $ \scriptsize{$\pm 80.945$} & $12.660 $\scriptsize{$\pm 2.084$} \\
 & & $2 {\times} 4 {\times} 2 {\times} 1$ & 2406 & $\mathbf{8.150}$ \scriptsize{$\pm 1.343$} & $\mathbf{1.811}$\scriptsize{$\pm 0.211$} \\
 \cline{2-6}
 & \multirow{2}{*}{FastKAN} & $4 {\times} 2 {\times} 2$ & 788 & $364.340 $ \scriptsize{$\pm 30.439$} & $13.180 $\scriptsize{$\pm 0.598$} \\
 &  & $4 {\times} 6 {\times} 2 {\times} 3 {\times} 2$ & 3155 & $442.129 $ \scriptsize{$\pm 54.164$} & $13.688 $\scriptsize{$\pm 0.707$} \\
 \cline{2-6}
 & \multirow{2}{*}{KAN} & $4 {\times} 2 {\times} 2$ & 876 & $508.330 $ \scriptsize{$\pm 67.823$} & $15.520 $\scriptsize{$\pm 0.902$} \\
 &  & $4 {\times} 6 {\times} 2 {\times} 3 {\times} 2$ & 3504 & $439.131 $ \scriptsize{$\pm 44.733$} & $12.932 $\scriptsize{$\pm 0.251$} \\
 \hline
\end{tabular}
    \label{tab:results_ff_dummy}
    \end{center}
\end{table*}
    
\subsection{Physical Equations}\label{sec:exp_physical}
For the physically meaningful formulae we use the same experimental setup as in subsection \ref{sec:exp_funcfit}. However, due to the higher complexity of this problem, we chose to generate datasets with $100,000$ samples instead to cover the input space more thoroughly. The first formula we used to generate a dataset originates from holography, where we have a reference beam with electric field strength $E_R$ and an object beam $E_0$. To reconstruct the holography image, a second reconstruction beam $\widehat{E}_R$ is used. The formula for the reconstructed hologram - which we want to learn - then states:
\begin{equation}
    H = \widehat{E_R} \cdot |E_R + E_0|^2
    \label{eq:holography}
\end{equation}
The second formula describes an alternating current electric circuit consisting of a resistor with resistance $R_L$ and a capacitor with impedance $\frac{1}{i\omega C}$ both in parallel to each other connected to a resistor $R_G$ and a coil with impedance $i\omega L$ in series. $R_L$ and $R_G$ are the resistances measured in ohm, $C$ is the capacitance of the capacitor measured in farad, $L$ is the inductance of the coil measured in henry and $\omega$ is the frequency of electricity.
We want to calculate the complex-valued voltage across the resistor $\underline{U}_{R_L}$ when the other quantities are given:
\begin{equation}
    \underline{U}_{R_L} = \frac{\underline{U}_G}{1 + \frac{R_G}{R_L}-\omega^2 \cdot L \cdot C + i \omega \cdot (\frac{L}{R_L} + R_G  \cdot C)}
    \label{eq:ac_circuit}
\end{equation}
Note that the features $R_G$, $R_L$, $L$, $C$ and $\omega$ are real-valued and only the two voltages $\underline{U}_{G}$ and $\underline{U}_{R_L}$ are complex-valued. However, the computation requires the use of complex-valued arithmetic.

The results of training on the holography dataset (cf. Table \ref{tab:physical_funcfit_holo}) show that models with more parameters perform better. While our \ac{ckan} has the best \ac{mse} score, FastKAN performs better than our approach on \ac{mae} with less trainable parameters. This hints that the \ac{ckan} is more stable against outliers, but trades off this stability for a less accurate average result.
When focusing on smaller - and thus more explainable - models, our model with size $3{\times}10{\times}1$ is the only small model to perform competitive with the best results. 

A similar picture can be observed for the circuit dataset (cf. Table \ref{tab:physical_funcfit_circuit}). In the \ac{mse} score, our \ac{ckan} outperforms other methods, however in the \ac{mae} score FastKAN performs better. Notably, on this dataset scaling up the model does not increase the performance meaningfully, with the best overall \ac{mae} score being obtained with the smallest FastKAN model. 

\begin{table}[tbp!]
    \begin{center}
    \caption{Results on the Holography Dataset \eqref{eq:holography}.}
    \addtolength{\tabcolsep}{-0.5em}  
\begin{tabular}{|c|c|c|c|c|}
\hline
Model & Size & \# Params & Test MSE & Test MAE \\\hline
\multirow{6}{*}{\acs{ckan}} & $3 {\times} 1$ & 396 & $53.020 $ \scriptsize{$\pm 0.393$} & $5.491 $\scriptsize{$\pm 0.019$} \\
 & $3 {\times} 1 {\times} 1$ & 533 & $41.889 $ \scriptsize{$\pm 0.222$} & $4.861 $\scriptsize{$\pm 0.017$} \\
 & $3 {\times} 3 {\times} 1$ & 1599 & $6.598 $ \scriptsize{$\pm 0.234$} & $2.038 $\scriptsize{$\pm 0.035$} \\
 & $3 {\times} 10 {\times} 1$ & 5330 & $0.151 $ \scriptsize{$\pm 0.006$} & $0.310 $\scriptsize{$\pm 0.011$} \\
 & $3 {\times} 10 {\times} 3 {\times} 1$ & 8381 & $0.168 $ \scriptsize{$\pm 0.043$} & $0.300 $\scriptsize{$\pm 0.037$} \\
 & $3 {\times} 10 {\times} 5 {\times} 3 {\times} 1$ & 13026 & $\mathbf{0.112}$ \scriptsize{$\pm 0.029$} & $0.240 $\scriptsize{$\pm 0.028$} \\
\hline
\multirow{4}{*}{FastKAN} & $6 {\times} 1 {\times} 2$ & 525 & $27.986 $ \scriptsize{$\pm 0.587$} & $3.687 $\scriptsize{$\pm 0.057$} \\
 & $6 {\times} 5 {\times} 2$ & 2617 & $7.640 $ \scriptsize{$\pm 2.454$} & $1.903 $\scriptsize{$\pm 0.306$} \\
 & $6 {\times} 10 {\times} 2$ & 5232 & $2.869 $ \scriptsize{$\pm 0.491$} & $1.182 $\scriptsize{$\pm 0.108$} \\
 & $6 {\times} 10 {\times} 5 {\times} 3 {\times} 2$ & 8571 & $0.140 $ \scriptsize{$\pm 0.102$} & $\mathbf{0.227}$\scriptsize{$\pm 0.077$} \\
\hline
\multirow{4}{*}{KAN} & $6 {\times} 1 {\times} 2$ & 584 & $38.295 $ \scriptsize{$\pm 6.309$} & $4.141 $\scriptsize{$\pm 0.282$} \\
 & $6 {\times} 5 {\times} 2$ & 2920 & $15.890 $ \scriptsize{$\pm 0.508$} & $2.745 $\scriptsize{$\pm 0.035$} \\
 & $6 {\times} 10 {\times} 2$ & 5840 & $1.853 $ \scriptsize{$\pm 1.096$} & $0.801 $\scriptsize{$\pm 0.341$} \\
 & $6 {\times} 10 {\times} 5 {\times} 3 {\times} 2$ & 9563 & $45.596 $ \scriptsize{$\pm 4.760$} & $4.757 $\scriptsize{$\pm 0.215$} \\
\hline
\end{tabular}
    \label{tab:physical_funcfit_holo}
    \end{center}
\end{table}

\begin{table}[tbp!]
    \begin{center}
    \caption{Results on the Circuit Dataset \eqref{eq:ac_circuit}.}
    \addtolength{\tabcolsep}{-0.5em}
\begin{tabular}{|c|c|c|c|c|}
\hline
Model & Size & \# Params & Test MSE & Test MAE \\\hline
\multirow{6}{*}{\acs{ckan}} & $6 {\times} 1$ & 792 & $7.166 $ \scriptsize{$\pm 1.343$} & $1.008 $\scriptsize{$\pm 0.010$} \\
 & $6 {\times} 1 {\times} 1$ & 929 & $6.840 $ \scriptsize{$\pm 1.362$} & $0.929 $\scriptsize{$\pm 0.009$} \\
 & $6 {\times} 3 {\times} 1$ & 2787 & $6.904 $ \scriptsize{$\pm 1.184$} & $0.938 $\scriptsize{$\pm 0.007$} \\
 & $6 {\times} 10 {\times} 1$ & 9290 & $\mathbf{6.780}$ \scriptsize{$\pm 1.458$} & $0.907 $\scriptsize{$\pm 0.009$} \\
 & $6 {\times} 10 {\times} 3 {\times} 1$ & 12341 & $8.101 $ \scriptsize{$\pm 1.044$} & $0.942 $\scriptsize{$\pm 0.025$} \\
 & $6 {\times} 10 {\times} 5 {\times} 3 {\times} 1$ & 16986 & $7.558 $ \scriptsize{$\pm 1.882$} & $0.798 $\scriptsize{$\pm 0.037$} \\
\hline
\multirow{4}{*}{FastKAN} & $7 {\times} 1 {\times} 2$ & 590 & $8.286 $ \scriptsize{$\pm 4.070$} & $\mathbf{0.708}$\scriptsize{$\pm 0.014$} \\
 & $7 {\times} 5 {\times} 2$ & 2942 & $10.236 $ \scriptsize{$\pm 3.712$} & $0.852 $\scriptsize{$\pm 0.011$} \\
 & $7 {\times} 10 {\times} 2$ & 5882 & $10.881 $ \scriptsize{$\pm 3.512$} & $0.986 $\scriptsize{$\pm 0.026$} \\
 & $7 {\times} 10 {\times} 5 {\times} 3 {\times} 2$ & 9221 & $8.799 $ \scriptsize{$\pm 4.025$} & $0.734 $\scriptsize{$\pm 0.027$} \\
\hline
\multirow{4}{*}{KAN} & $7 {\times} 1 {\times} 2$ & 657 & $12.984 $ \scriptsize{$\pm 12.854$} & $0.764 $\scriptsize{$\pm 0.111$} \\
 & $7 {\times} 5 {\times} 2$ & 3285 & $9.051 $ \scriptsize{$\pm 4.766$} & $0.864 $\scriptsize{$\pm 0.019$} \\
 & $7 {\times} 10 {\times} 2$ & 6570 & $8.911 $ \scriptsize{$\pm 4.119$} & $1.015 $\scriptsize{$\pm 0.018$} \\
 & $7 {\times} 10 {\times} 5 {\times} 3 {\times} 2$ & 10293 & $8.840 $ \scriptsize{$\pm 3.981$} & $0.990 $\scriptsize{$\pm 0.019$} \\
\hline
\end{tabular}
    \label{tab:physical_funcfit_circuit}
    \end{center}
\end{table}

\subsection{Knot Classification}
\label{sec:results_knot}
The knot dataset \cite{davies2021advancing} originates from knot theory. Davies et al. calculated 15 invariants on $\approx240.000$ different knots, from which 13 are real-and two complex-valued. Thus we need a real-valued \ac{kan} with 17 input features or a \ac{ckan} with 15 complex-valued input features, out of which 13 have their imaginary part set to zero. The task on this dataset is to classify the knots based on their invariants into 14 different classes representing their signature. Therefore the output layer's width of both real- and complex-valued \acp{kan} is set to $14$ to produce a probability vector. \done{$\C\to\R$ \ac{ckan}}
Since these probabilities are real-valued we need to learn a function $\C\to\R$ in the output layer as described in subsection \ref{sec:ckan_crbf}. 
The features of the dataset are normalized to fit in our grid of fixed size $[-2,2]$ for real-valued \acp{kan} and $[(-2-2i), (2+2i)]$ for \ac{ckan}.

\begin{figure}[t]
    \begin{center}  
    \includegraphics[width=\linewidth]{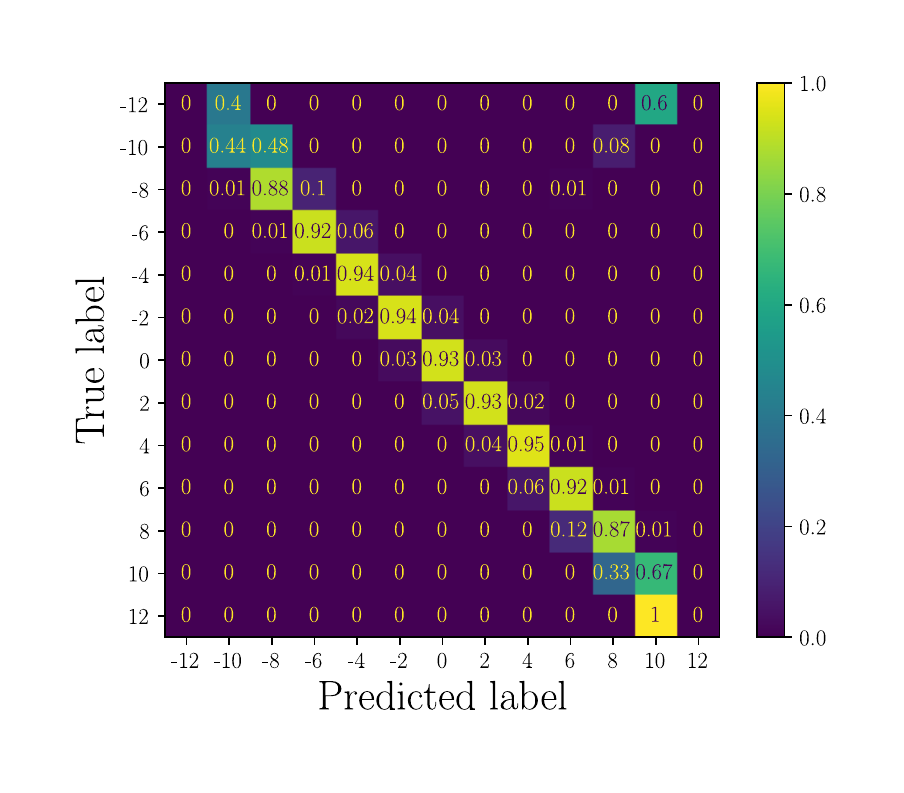}
    \vspace{-8mm}
    \caption{Confusion matrix for \ac{ckan} on the knot dataset normalized to probabilities within each row.}
    \label{fig:ckan_knot_confusionmatrix}
    \end{center}
\end{figure}

In Table \ref{tab:knot_summary} we list the results of \acs{kan}, FastKAN and \acs{ckan} for hidden layer widths $\in \{1,2\}$ and a grid size of $G=8$. For the two real-valued \acp{kan} we also experimented with a grid size of $G=64$ each to conduct a fairer comparison of our \acs{ckan} to the others. Our approach is the best-performing model on the knot dataset. While the best \ac{ckan} also has the highest number of parameters, the $15{\times}1{\times}14$ \ac{ckan} performs still better than FastKAN and approximately equally well as \ac{kan} with a similar amount of parameters.
In Fig. \ref{fig:ckan_knot_confusionmatrix} an exemplary confusion matrix of our best performing \ac{ckan} is shown. The classes -12 and 12 are highly underrepresented in the dataset and thus \ac{ckan} does not learn anything for these classes. Besides these outliers, all classes are learned well.

Fig. \ref{fig:ckan_knot_relevances} shows the first layer of a \ac{ckan} trained on the knot dataset with the opacity of nodes and edges showcasing the calculated relevance scores. Like in \cite{liu2024kan} the meridinal, here as one single complex number, and longitudinal translation are the most relevant features.
To also quantitatively evaluate the explainability, we conduct a small study, where we exclude less relevant features and retrain the model on only the more relevant features. When reducing the dataset with this method to 7 or 3 features, we still obtain good results, only reducing the accuracy by $0.012$ and $0.028$ respectively (cf. Table \ref{tab:knot_ckan_reducedtrainfeatures}). 
In contrast, when we leave out these 7 or 3 most relevant features, the performance drops to $\approx 30\%$. This shows that this method determines the relevance of features accurately.
\begin{table}[tbp!]
    \begin{center}
    \caption{The best-performing \acp{ckan} for hidden layer widths $\in \{1,2\}$, all experiments with FastKAN and \ac{kan} on the knot dataset.}
\begin{tabular}{|c|c|c|c|c|}
\hline
Model & Size & \#Params & Test Acc. & Test CE-Loss \\\hline
\multirow{2}{*}{\acs{ckan}} & $15{\times}1{\times}14$ & 2921 & $0.923 \pm 0.001$ & $0.212 \pm 0.001$ \\
& $15 {\times} 2 {\times} 14$ & 6754 & $\mathbf{0.944}$ \scriptsize{$\pm 0.001$ }& $\mathbf{0.151}$ \scriptsize{$\pm 0.003$} \\
\hline
\multirow{4}{*}{FastKAN} & $17 {\times} 1 {\times} 14$ & 296 & $0.739$ \scriptsize{$ \pm 0.007$} & $0.609 $ \scriptsize{$\pm 0.005$} \\
& $17 {\times} 1 {\times} 14$ & 2032 & $0.893 $ \scriptsize{$\pm 0.003$} & $0.331 $ \scriptsize{$\pm 0.013$} \\
 & $17 {\times} 2 {\times} 14$ & 578 & $0.898 $ \scriptsize{$\pm 0.002$} & $0.259 $ \scriptsize{$\pm 0.002$} \\
 & $17 {\times} 2 {\times} 14$ & 4050 &  $0.894 $ \scriptsize{$\pm 0.016$} & $0.315 $ \scriptsize{$\pm 0.052$} \\
\hline
\multirow{4}{*}{KAN} & $17 {\times} 1 {\times} 14$ & 527 & $0.835 $ \scriptsize{$\pm 0.009$} & $0.442 $ \scriptsize{$\pm 0.012$} \\
 & $17 {\times} 1 {\times} 14$ & 2263 & $0.881 $ \scriptsize{$\pm 0.007$} & $0.620 $ \scriptsize{$\pm 0.520$} \\
 & $17 {\times} 2 {\times} 14$ & 1054 & $0.903 $ \scriptsize{$\pm 0.033$} & $0.282 $ \scriptsize{$\pm 0.123$} \\
 & $17 {\times} 2 {\times} 14$ & 4526 & $0.938 $ \scriptsize{$\pm 0.001$} & $0.184 $ \scriptsize{$\pm 0.004$} \\
\hline
\end{tabular}
    \label{tab:knot_summary}
    \end{center}
\end{table}

\subsection{Ablation Study}
The ablation study regarding the influence of the used normalization scheme (cf. Section \ref{sec:cvkan_cvbn}) and the choice of weights for \acs{csilu} (cf. section \ref{sec:cvkan_residual})  was conducted for the knot dataset and not for the synthetic function fitting datasets. We found in Table \ref{tab:knot_ckan_ablation} that normalization has a positive impact on the model but the choice of normalization scheme has little importance as $\text{BN}_\C$, $\text{BN}_\mathbb{V}$ and $\text{BN}_{\R^2}$ perform all similarly well but better than no normalization. As for the choice of \acs{csilu} it shows that the complex-weighted variant $\text{\acs{csilu}}_\C$ is always just as good or better than the real-weighted $\text{\acs{csilu}}_\R$.

\begin{table}[tbp!]
    \begin{center}
    \caption{Results of training a \ac{ckan} (with one intermediate layer of size 1) on the full knot dataset ($15$ features), and using only the most relevant $7$ or $3$ features or leaving these out.}
\begin{tabular}{|c|c|c|c|c|}
\hline
Features & \# Params & Test Acc. & Test CE-Loss\\
\hline
all  15 & 2921 & $0.923$ \scriptsize{$\pm 0.001$} & $0.212$ \scriptsize{$\pm 0.001$}\\
\hline
only $7$ most important & 1867 & $0.911$ \scriptsize{$\pm 0.003$} & $0.233$ \scriptsize{$\pm 0.003$}\\
only $3$ most important & 1339 & $0.895$ \scriptsize{$\pm 0.002$} & $0.267$ \scriptsize{$\pm 0.002$}\\
\hline
all but $7$ most important & 1999 & $0.284$ \scriptsize{$\pm 0.002$} & $1.757$ \scriptsize{$\pm 0.003$}\\
all but $3$ most important & 2527 & $0.300$ \scriptsize{$\pm 0.003$} & $1.663$ \scriptsize{$\pm 0.002$}\\
\hline
\end{tabular}
    \label{tab:knot_ckan_reducedtrainfeatures}
    \end{center}
\end{table}

\begin{table}[tbp!]
    \begin{center}
    \caption{Ablation study on the influence of normalization scheme and \acs{csilu} type.  Evaluated on a $15{\times}1{\times}14$ \ac{ckan} on the knot dataset.}
\begin{tabular}{|c|c|c|c|c|}
\hline
\# Params & Normalization & \acs{csilu} & Test Acc. & Test CE-Loss \\\hline
\multirow{2}{*}{2923} & \multirow{2}{*}{$\text{BN}_\C$} & c & $0.921 $\scriptsize{$\pm 0.001$} & $0.213 $\scriptsize{$\pm 0.001$} \\
 & & r & $0.921 $\scriptsize{$\pm 0.002$} & $0.213 $\scriptsize{$\pm 0.001$} \\
\hline
\multirow{2}{*}{2921} & \multirow{2}{*}{$\text{BN}_\mathbb{V}$} & c & $\mathbf{0.923}$\scriptsize{$\pm 0.001$} & $\mathbf{0.212}$\scriptsize{$\pm 0.001$} \\
 & & r & $0.921 $\scriptsize{$\pm 0.001$} & $0.213 $\scriptsize{$\pm 0.001$} \\
\hline
\multirow{2}{*}{2922} & \multirow{2}{*}{$\BN_{\R^2}$} & c & $0.920 $\scriptsize{$\pm 0.001$} & $0.217 $\scriptsize{$\pm 0.001$} \\
 & & r & $0.920 $\scriptsize{$\pm 0.001$} & $0.213 $\scriptsize{$\pm 0.001$} \\
\hline
\multirow{2}{*}{2918} & \multirow{2}{*}{none} & c & $0.886 $\scriptsize{$\pm 0.016$} & $0.317 $\scriptsize{$\pm 0.046$} \\
 & & r & $0.789 $\scriptsize{$\pm 0.153$} & $0.591 $\scriptsize{$\pm 0.445$} \\
\hline

\end{tabular}
    \label{tab:knot_ckan_ablation}
    \end{center}
\end{table}

\section{Discussion}
\done{filled and reworked discusion}
We have shown that \ac{ckan} has a clear advantage over the real-valued \acp{kan} when dealing with complex-valued function-fitting tasks. Our approach can compete with \ac{kan} and FastKAN while having less parameters and a shallower network architecture. These properties affect the explainability in a positive way. Furthermore our \ac{ckan} has proven to be more stable w.r.t. outliers in regression tasks since \ac{ckan} always produces one of the best \ac{mse} scores but is sometimes outperformed on the \ac{mae} metric. Our approach also produces more stable results across the five different runs of cross-validation for each configuration, which underlines the consistently good generalization capabilities of our model.

While \ac{ckan} does not excel on datasets with mostly real-valued features, like the circuit dataset in subsection \ref{sec:exp_physical}, it still performs similarly well as the real-valued \acp{kan}. If the datasets contain more complex numbers, \ac{ckan} can unfold it's full potential and outperform the real-valued \acp{kan} while requiring less parameters and having a shallower network structure, which improves the explainability.

Our tool for visualizing the proposed \ac{ckan} enables the understanding of the calculations happening inside our \ac{ckan} and can thus help with (re-) discovering mathematical correlations in the data like Liu et al. pointed out in their real-valued \ac{kan} \cite{liu2024kan, liu2024kan20}. The calculated importance scores reflect the true relevances of the features, as is shown by our experiments of training only on the $3$ or $7$ most important features or by leaving them out in Table \ref{tab:knot_ckan_reducedtrainfeatures}.
While the plot of our \ac{ckan} trained on synthetic formula $f_4$ in Fig. \ref{fig:ckan_sqsq} may seem unintuitive at first glance, the learned functions that are plotted show an interesting insight. In the bottom layer there exist two roots in both functions for $z_1^2$ and $z_2^2$. The root points are exactly rotated around $90$° when comparing those two functions. This is caused by shifts that cancel out: 
\begin{equation}
(z_1^2 + s) + (z_2^2 -s)  = z_1 ^2 + z_2 ^2
\end{equation}
Therefore we end up with roots at $\pm \sqrt{s}$ or $\pm \sqrt{-s}$ in the respective functions, which are identical roots up to $\sqrt{-1}=i$, which corresponds to a rotation of 90° in the complex plane. 

Besides the qualitative visualization, the calculated relevance scores on the edges allow to analyze the importance of input as well as intermediate features. We show that this leads to meaningful results in our case study (cf. Table \ref{tab:knot_ckan_reducedtrainfeatures}). 

However, further research is required to study the actual explainability of \acp{kan} on real-world datasets.

\section{Conclusion}
In this work we have developed \ac{ckan}, a complex-valued variant of \ac{kan}, which relies on complex-valued \ac{rbf}s. We have introduced BatchNorm to efficiently tackle the problem of outputs exceeding the grid size of subsequent layers. In our experiment we could show that \ac{ckan} outperforms the real-valued versions \ac{kan} and FastKAN consistently on complex-valued tasks and performs on par or superior on tasks that mix complex- and real-valued inputs. Additionally, \ac{ckan} training is more stable and leads to more explainable models, since less layers are needed to produce good results. 

As future work, we aim to apply \ac{ckan} to real world problems with complex-valued data, such as finding solutions to complex-valued PDEs by utilizing them in Kolmogorov-Arnold-Informed Neural Networks \cite{wang2025kolmogorov}. Additionally, the concept of \acs{ckan} could be extended further to other arbitrary hypercomplex algebras such as quaternions \cite{parcollet2020survey,lopez2024towards}.
\bibliographystyle{IEEEtran}
\bibliography{References}
\end{document}